# Modified SMOTE Using Mutual Information and Different Sorts of Entropies

Sima Sharifirad, Azra Nazari, Mehdi Ghatee*

*Abstract*—SMOTE is one of the oversampling techniques for balancing the datasets and it is considered as a pre-processing step in learning algorithms. In this paper, four new enhanced SMOTE are proposed that include an improved version of KNN in which the attribute weights are defined by mutual information firstly and then they are replaced by maximum entropy, Renyi entropy and Tsallis entropy. These four pre-processing methods are combined with 1NN and J48 classifiers and their performance are compared with the previous methods on 11 imbalanced datasets from KEEL repository. The results show that these pre-processing methods improves the accuracy compared with the previous stablished works. In addition, as a case study, the first pre-processing method is applied on transportation data of Tehran-Bazargan Highway in Iran with IR equal to 36.

## I. Introduction

One of the topics which attracts a lot of attention in machine learning is imbalanced data. In fact, in real world the data are imbalanced. The class imbalanced problem has wide applications from telecommunication [1], identification of oil spoils in satellite radar images [2], medical diagnosis [3], financial fraud detection [4], network intrusion detection [5], text classification [6] and so forth. In all described usages, the minority class has high sensitivity and its information is the target. However, classification algorithms are attracted to the majority class and simply ignore the minority one. Thus, they are mostly producing poor accuracy prediction over the minority class.

Generally, a dataset is considered imbalanced when the class distribution is unequal and the number of one class is outnumbered the other one. Practically, the problem of unequal proportion between classes happens when the learning classifier makes decisions biased toward the majority class [10]. Moreover, reaching high accuracy in data mining is very important [7]. In the context of two class imbalanced problems, the majority class is referred to as negative while the minority class as positive. From one point of view, managing imbalanced data to reach the best performance is the target of both training and evaluating phase. Moreover, intrinsic datasets referred to data space while the extrinsic imbalanced datasets are not related to data space. Generally, three categories of solutions are proposed namely data-level or external level, algorithmic-level or internal –level and cost-sensitive approaches [8].

In the data level approach, many pre-processing methods are proposed that try to balance the data before testing with different classifiers. These methods are mainly divided into three categories, namely, oversampling, under sampling and hybrid methods. In algorithmic level, new classification algorithms are designed or modified to be suitable for imbalanced data. In addition, cost-sensitive approaches have been proposed to control the problem of different misclassification error costs which may be used for imbalanced datasets [9].

The rest of this paper is organized as the following: Section 2 includes some preliminaries. In Section 3, a concise description of new methods is given. Section 4 consists of simulation results. Final section summarizes the paper with a brief conclusion.

## II. Preliminaries

### A. 2.1 Data level

In the data level phase, the class distribution is modified on the training procedure by producing new samples randomly or artificially or by omitting non-informative samples [11].

The first category is over sampling. This method is firstly proposed by [12] and mainly generates positive samples. In fact, samples of minority class are replicated specific times. For instance, 100% oversampling means that the minority class samples are duplicated once and minority instances are doubled [14]. In terms of time complexity, this method has high efficiency when managing a large volume of data [15]. Beside the mentioned privileges, it has some shortcomings too.

This method decreases the decision region more specifically and as a result leads to over fitting. One of its kind is random oversampling that randomly selects the examples from the minority class and replace them until the number of chosen samples plus the original examples of the minority class becomes equal to the majority one [16,46]. In other word, this method is considered as a non-heuristic method that balances the distribution of class by random replication [17, 18]. One of the most famous kinds of oversampling is synthetic minority oversampling which produces artificial samples.

The authors are with the Department of Computer Science, Amirkabir University of Technology, Tehran, Iran. *The corresponding author (Email: ghatee@aut.ac.ir))

The second category is considered as under sampling, the simplest method of under sampling is random under sampling (RUS) which omitted the samples of the majority class randomly until their numbers matched the number of minority class samples [19,46]. Its merit is because of the time which is required to train models is reduced due to the fact that the training datasets are reduced [20]. Along with the mentioned merits, it has demerits too. Its first problem is that in most cases the useful information of the majority class is discarded. The second related problem is that, in real world application like bankruptcy prediction task this method is not practical [21].

The third category is hybrid method which is first proposed by [22]. This method combines the two previous methods which are oversampling and under sampling. In this method, the merits of the methods combined while their drawbacks are decreased dramatically.

### B. SMOTE and its kind

Smote is first proposed by [23]. This method produces synthetic samples of the minority class by working in the "feature space" rather than "data space" [25]. Generally, classifiers are allowed to predict unseen examples belonging to the minority class and learned more general regions for the minority class [26]. Practically, this method generates artificial members from the minority class by multiplying each feature of original object times a random number between 0 and 1 and adding up this result to the original features.

SMOTEboost is proposed as a new method which mixes standard boosting process with SMOTE. This method lets learner sample more from the minority class and make decision region broader [26]. It produces artificial samples located on the path between two neighbors from the minority class [27]. Borderline-smote is another modification by [28]. This method works in the procedure of sampling and locates the minority data points closer to the separation boundary. It first finds the minority class samples near to the separation boundary and call it "Danger" region and then use smote to produce synthetic objects for each instance in the "danger" set [29]. Safe-level SMOTE is another method which carefully chooses instances from minority instances along the same line with different weights which is called safe level. This method before producing synthetic samples allocates each synthetic samples to a specific safe level [30]. Each synthetic sample located closer to the largest safe level so all instanced are produced only in safe regions. Other representative works includes the OSS method [31], integration method [32], the neighborhood cleaning rule (NCL) [33].

### C. KNN and its Extensions

K nearest neighbor is first proposed in [34] and is considered as one of the top ten algorithms. Because of its high computational time and cost during classification time, the model is not built during training period. Its positive features like simplicity, ease of implementation and its effectiveness leads to its extensive use in different areas. Nevertheless, this algorithm suffers from some deficits which have direct impact on its accuracy.

Different research has been conducted on this algorithm, one of them which is related to the selection of k is selective neighborhood naïve Bayes that different values of k are assessed and naïve Bayes is trained for each of them [35]. The other one is dynamic KNN that combines two methods of eager and lazy learning [36]. In the simple form of KNN, the Euclidean distance is measured based on the equal distribution of all attributes while in weighted KNN, each attribute has a degree of importance.

### D. Different sort of entropies

Entropy has a wide range of definitions in different fields such as thermodynamic, statistics, disorder, information and homogeneity context. In 1948, it is first proposed in information theory context by Shannon [38]. The most important entropies are defined as the following:

- Renyi entropy

Which is considered as a generalized form of Shannon entropy is first proposed by Renyi [43]. It has flexibility in terms of parameter α which shows order and p which is a probability distribution on a finite set. [39]. in this research Renyi entropy with order 2 is used which is called collision entropy [41]. Its original formula is as the following:

$$I_\alpha(p) = 1/(1-\alpha) \log \left( \sum_{k=1}^{N} p^\alpha \right) \qquad (1)$$

- maximum entropy

This method [42] is considered as either a usual method of inference of classical statistics or representative of their conceptual generalization. Its aim is to maximize the information entropy. Moreover, the estimation of the probability distributions from data are estimated by this general technique. Its overriding principle is when nothing is known and the distribution should be as uniform as possible. In addition, it is unique and agrees with the maximum-likelihood distribution and has the exponential form [41]. Its formula is as the following:

$$S = \sum P(A) \log 1/(P(A)) \qquad (2)$$

- Tsallis entropy

This is a generalized form of Boltzmann-Gibbs entropy and is proposed in 1988 due to the fact that it has no extensive physical system. Its application is from natural, artificial and social complex systems. Based on Google scholar, since its introduction, it is cited for 5314 times. This entropy is firstly used in decision tree and reach high accuracy. It generally experiences good performance on data which has large number of features and very small number of samples [42]. Its formula is:

$$S_\alpha = 1/(\alpha - 1)\left(1 - \sum_{i=1}^{N} p_i^\alpha\right) \qquad (3)$$

- Mutual information

This is one of the methods that tries to find how much uncertainty can be omitted in one model by the state of the next variable [43]. The arbitrary dependencies between random variables are measured and are used for evaluating the features in complex classification tasks based on linear relations. This method usually decreases the initial uncertainty, in the ideal case, this amount is usually set to zero.

In this article we use different sorts of entropy as a given weight for our features. Firstly, we use mutual information as a weight, as declared before, it uses Shannon entropy as one of the factors. Later we use three other entropies instead of mutual information. It is worth mentioning that for more clarity in this article, we use abbreviation form of methods. Instead of mutual information enhanced SMOTE, we used MISMOTE, maximum enhanced entropy SMOTE as MAESMOTE, Tsallis enhanced entropy SMOTE as TESMOTE and Renyi entropy enhanced SMOTE as RESMOTE.

*E. Evaluation metrics*

Generally, for evaluating the performance of a machine learning algorithm a confusion matrix as shown in Table 1 is used for rare class problems. In classification issues, assuming class "C" as the minority class while "NC" is a conjunction of all the other classes. Usually there are four possible results from class "C".

| Table1. Confusion matrix | | |
|---|---|---|
| | Predicted class "C" | Predicted class "NC" |
| Actual class "C" | True positive(TP) | False negative(FN) |
| Actual class "NC" | False positive(FP) | True negative(TN) |

Some other metrics like recall, precision and f-value can be described as follows:

Precision=$TP/((TP+FP))$ (4)

Recall=$TP/((TP+FN))$ (5)

$$F\text{-value} = \frac{(1+\beta^2).\text{recall}.\text{precision}}{\beta^2.\text{recall}+\text{precision}} \quad (6)$$

In the above formula, $\beta$ is related to relative importance of precision vs. recall and it is usually supposed to be 1. Although, the main goal of all classification algorithms are enhancing recall without decreasing the precision, in most cases, precision and recall are often opposing to each other since by the increased value of true positive, false positive is increased and as a result the precision reduces. Moreover, when one class is rare they may not work well enough simultaneously. The f-value considers both precision and recall, shows the relative impact of precision and recall by one number. They are used to evaluate the performance of algorithm in terms of minority class [25]. To sum up, when both values of precision and recall are large, f-value is large, too.

Another evaluation metric which is used for comparing the imbalanced algorithms is AUC which is area under ROC (receiver operating characteristic) curve and is first proposed by Bradley, 1997. In fact, for summarizing the performance of the algorithms, this measure is used as a standard technique between TP rate (benefits) and FP rate (cost). To calculate the AUC, we need to have the area of the graph as following [43]:

$A_{ROC} = (1+TP_{rate} - FP_{rate})/2$ (7)

### III. PROPOSED PRE-PROCESSING METHODS

In this article we enhance the performance of SMOTE by replacing its KNN algorithm with the enhanced one along with the usage of different entropies and mutual information as the considered weight of features. As one can see in Fig. 1, due to the fact that our datasets didn't have any label, we use Hierarchical Agglomerative clustering algorithm and identify the labels. Then, we used mutual information and different sorts of entropies like Renyi entropy, Tsallis and maximum entropy each time and considered them as the weight for each feature. After that for identifying the appropriate K for KNN algorithm, we used leave-one-out cross validation. What is more, in the process of classification, the distance between the test sample and the mean of each cluster is identified, among them the closest is chosen and weighted Euclidian distance is carried between the sample test and each of the sample in that class. The first four phases of enhanced KNN which is considered as pre-processing, are done just once so they don't have a lot of effect. It is worth mentioning that these steps were fixed between all of our four methods. In the first step of implementation of SMOTE, N which is the amount of over-sampling is considered, its minimum amount is 100. Take an example that the percentage of N is 200%, in this case just two of the nearest neighbors are considered and one sample is generated in each direction. Flowchart.1 shows all the process of implementation step by step. In this flowchart, the first and the last steps are related to SMOTE and in the middle, as mentioned before KNN is enhanced by different sorts of methods such as dynamic selection, attribute weighted which is done initially by mutual information and then replaced by other entropies and distance weighted techniques. Information gain which is a weight coefficient of each attribute is calculated as the following [44]:

$$E(S) = -\sum_i p_i \log(p_i) \quad (8)$$

Where $p_i = |C_i, S|/|S|$ in this formula $p_i$ is the probability of random tuple in S and it belongs to class Ci. [44].

$$E(S|A) = -\sum_j (|S_j|/|S|)E(S_j) \quad (9)$$

In the above formula, A is a feature that has different values.

Gain (A) =E(S)-E (S│A) ≥0 (10)

$\lambda_i$=Gain (i)*$e^{Gain(i)}/\sum_{i=1}^{n} Gain(i)$ (11)

In this paper, for each attribute based on formula 8 we calculate Shannon entropy, after calculating their conditional entropy based on formula 9, we place them in formula 10 to calculate information gain. Then, a final weight is allocated for each feature based on formula 11 [45].

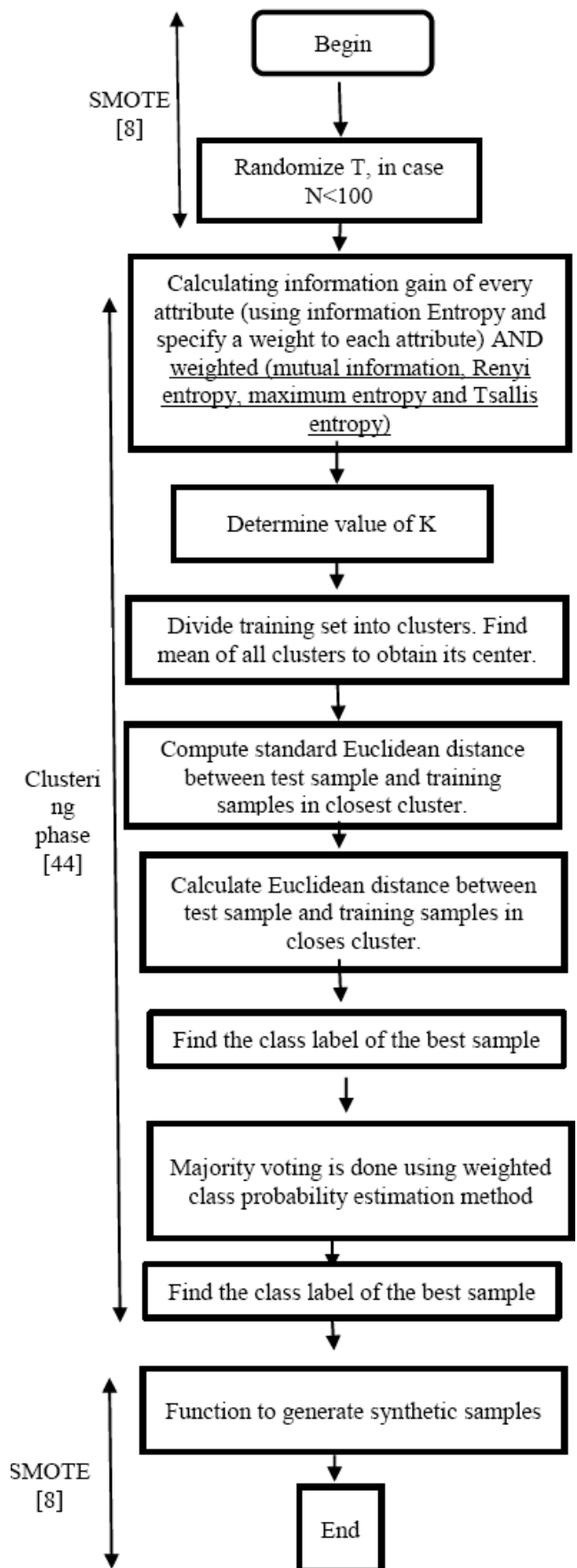

Fig. 1. The flowchart of the proposed system

## IV. SIMULATION RESULTS

The four proposed pre-processing methods have been applied on 11 imbalanced datasets with various distribution and imbalanced ratio(IR) namely Wisconsin, Ecoli1, Ecoli2, Ecoli3, Glass1, Glass6, Iris, Pima , newthyroid1 and yeast1 which were taken from KEEL(knowledge extraction based on evolutionary learning) dataset repository (http://www.keel.es/dataset.php), and a dataset which is related to the number of accidents and injuries that happened in Tehran-Bazargan Highway with 900 kilometer length and it is considered as one of the most sensible routes in terms of its business directory, economic and length in Iran.

All the implementation that leads to proposing the new methods for balancing datasets are implemented in excel and a macro is designed for each of the methods. In this way, the number of minority samples are imported to excel and then four macros based on each of our methods were implemented for producing synthetic samples in a new sheet. After getting the new datasets, they are imported in WEKA software to be checked with other classification algorithms. It is worth mentioning that all the datasets have been firstly changed into two class problems. Table 2 in appendix shows the main features of the datasets including the imbalanced ratio (IR), which is the number of negative samples divided by the number of positive samples. Data are gathered by police sources from 2010 to 2013. This data include 18 features that mainly are related to three types of accidents namely: injuries, damage and fatalities. In this model, we assume the most important categories as injuries and fatalities which follow imbalanced pattern.

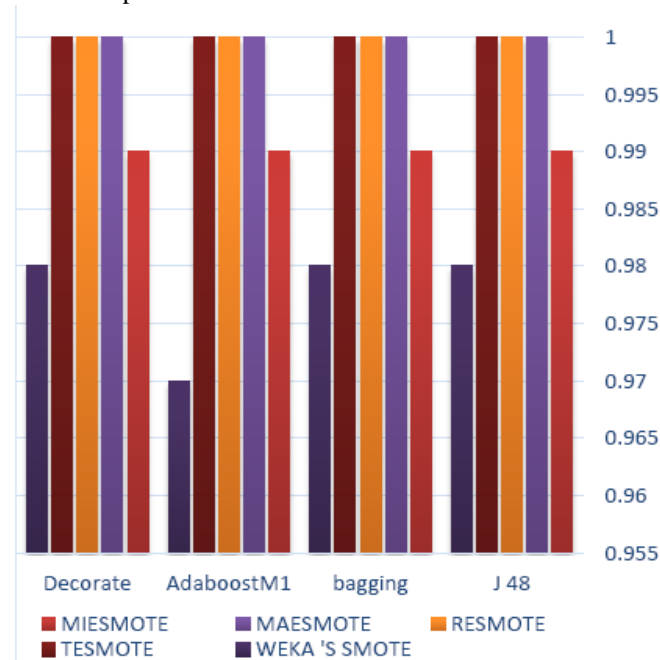

Fig. 2. Comparison between enhanced SMOTE and WEKA' s SMOTE on Tehran-Bazargan accident dataset

We choose injuries and fatalities that reveal the characteristic of imbalanced data more in term of their imbalance ratio (IR). The number of injuries are approximately about 24195 while fatalities 676 in one year.

As it is clear the number of injuries far outnumbered the number of fatalities by IR about 35.79. (http://www.keel.es/dataset.php).

Fig. 2 shows a comparison between WEKA's SMOTE and our four proposed methods. They are all considered as pre-processing methods. In order to reach an exact result, we compare these methods by four classification algorithms namely Decorate, AdaboostM1, Bagging and J48. In all the results, MAESMOTE, RESMOTE and TESMOTE have the same level of goodness and are considerably higher than MIESMOTE and WEKA's SMOTE. Meanwhile, MIESMOTE exhibits better performance in comparison to WEKA's SMOTE. It is clear from the data that there are some significant differences in the performance of our methods comparing to WEKA'S SMOTE.

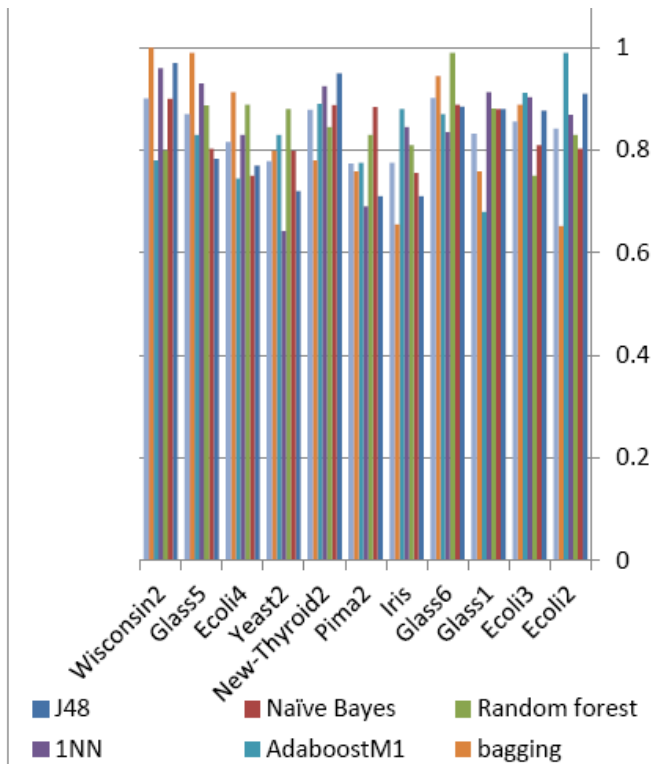

Fig. 3. AUC Comparison of 11 selected Keel datasets with MIESMOTE and 6 classification algorithms

Fig. 3. Shows comparison of AUC results for 11 selected KEEL datasets. MIESMOTE is considered as a pre-processing method and six classification algorithms from different categories are implemented for comparing the results. The first dataset, Wisconsin2 and Ecoli 3 have their best performance with bagging. The same trend comes true for Glass5, Ecoli4 and Glass6. Meanwhile, Yeast2 experiences its best performance with random forest. Specifically, Iris and Ecoli2 have their best results with AdaboostM1. While Pima2 acquires its best performance with Naïve Bayes, Glass1 keeps its highest score with one nearest neighbor and New-Thyroid2 with J48. It is clear from the data that there are significant differences in terms of different datasets and different classification algorithms.

In contrast, New-Thyriod2 and Ecoli2 have their minimum performance with Bagging algorithm. As for Iris and Ecoli3 have their worst performance with Random forest. Wisconsin2 and Glass1 experience their worst performance with AdaboostM1. It is noteworthy that Pima2, Glass6 and Yeast2 have their lowest score with one nearest neighbor while Glass5 with J48 and Ecoli4 with Naive Bayes.

As Fig.3 shows, among all of the 7 algorithms, AdaboostM1 provides the highest performance on Ecoli2, Iris, New-Thyroid, Yeast2. The worst performance obtained on Glass1, Ecoli4 and Winconsin2. In the case of Naïve Bayes, Ecoli4, and Pima2 have their highest score while Glass5 has its minimum performance. Meanwhile, Bagging with Ecoli3 and Glass5 obtains its peak but New-Thyroid2 reaches its minimum. As seen in the chart, Random forest is almost among the worst algorithms with Ecoli2, Ecoli3 and Iris. On the other hand, Glass6 and Wisconsin2 have their highest score. In comparison, one nearest neighbor has its rise with Glass1 and Iris but its minimum score with Pima2, Glass6 and Yeast2. Finally, J48 is the only algorithm that gives a minimum performance with Glass6.

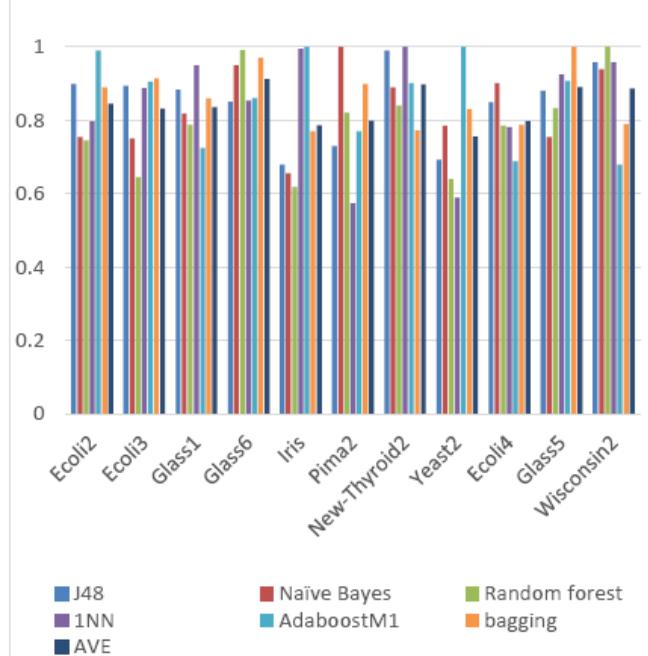

Fig 4. AUC Comparison of 11 selected Keel datasets with MAESMOTE and 6 classification algorithms.

In Fig.4, Ecoli2 corresponds to the best performance with Naïve Bayes but J48 has its minimum score with Random forest. On the other hand, Ecoli3, with one nearest neighbor reaches its peak while has the lowest score with random forest. Bagging has the best performance with Glass1, New-Thyroid2, Iris and Glass5. Meanwhile, Glass6 experiences highest level of performance with random forest and is minimum one with AdaboostM1. Moreover, Pima2 reaches the maximum result with AdaboostM1 and its minimum with J48. Random forest undergoes its least level of performance with Iris, New-Thyroid2 and Wisconsin2. Yeast 2 and Ecoli4 experience a summit with AdaboostM1 and a bottom level with Naïve Bayes. The last two datasets, Glass1 and Glass5 reach their lowest level of performance with J48 and

AdaboostM1 respectively.

Fig.5 demonstrates that Ecoli2, Iris, New0Thyroid2 and Yeast2 provide the best results with AdaboostM1 while their worst results obtained by Bagging, J48, 1NN and Yeast2 respectively. Meanwhile, Ecoli3 and Wisconsin2 reach their peak with J48 and their lowest point with random forest and Naïve Bayes. As far as Glass6, Pima2, Ecoli4 and Glass 5 have their best performance with random forest, their lowest score of the first two datasets are related to J48, and the second two datasets are related to Bagging and Naïve Bayes. Once again, Glass1 experiences its highest score with one nearest neighbor and its highest score with J48.

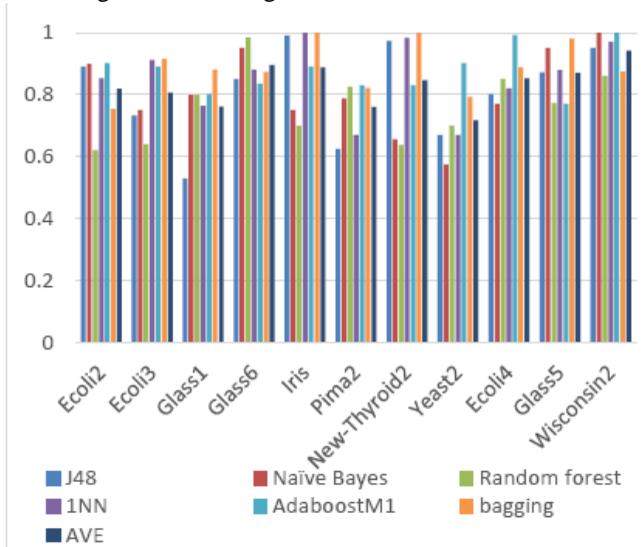

Fig. 5. AUC Comparison of 11 selected Keel datasets with REESMOTE and 6 classification algorithms.

Due to the fact that in [42] the best results are compared by two main algorithms namely 1NN and J48, Fig.6 and Fig.7 are presented for comparing the results with these classifiers. Also details of comparisons are given in Table 3 and Table 4 in the appendix.

As Fig. 6 illustrates, AdabostM1 is among the best classifiers when it is used with Ecoli2, Yeast2 and Iris. In addition, Random forest has its best performance with Glass6, Pima2, Ecoli4 and Glass5. Thirdly, some datasets such as Ecoli3 and Wisconsin experience their best performance with J48. Moreover, Glass1 and New-Thyroid2 with 1NN have their maximum performance.

Also Fig. 7 illustrates comparison between the best previous methods and our four proposed methods applying 1NN classifier. It shows that with Wisconsin2, TSESMOTE performs as good as the best previous methods but better than the previous ones. Unfortunately for Glass5, none of our methods conquer the best previous ones. In contrast, Ecoli4 with RESMOTE has the best performance. Meanwhile, with Yeast2, TESMOTE has better result with 0.02% improvement in comparison with the previous one and reaches 0.67. Moreover, with New-Thyroid2, MAESMOTE, RESMOTE and TSESMOTE obtain better results respectively. Moreover, with Pima2, MISMOTE reaches better performance. TSESMOTE has equal improvement in comparison to the best previous result in Iris. At the same time, with Glass6, RESMOTE has better performance in comparison to the best previous ones. Glass1 with MAESMOTE has better performance in comparison with the best previous methods. Last but not least, TESMOTE has an equal importance with other previous methods.

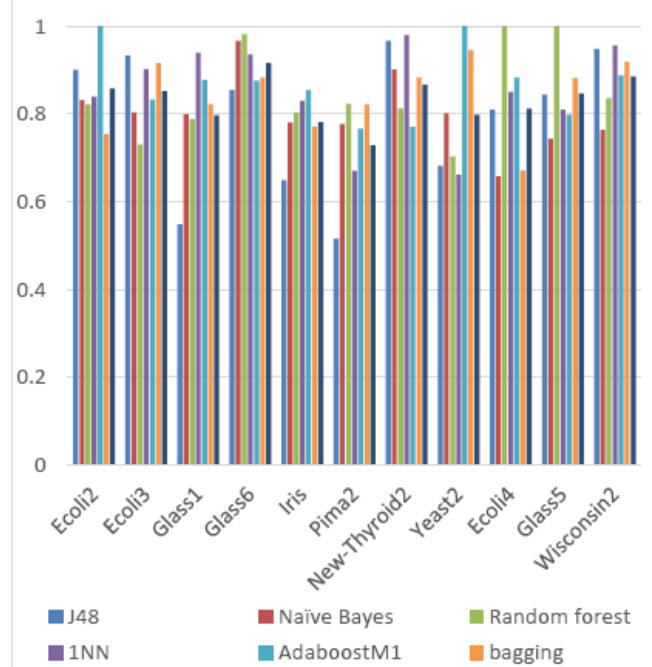

Fig. 6. AUC Comparison of 11 selected Keel datasets with TESMOTE and 6 classification algorithms.

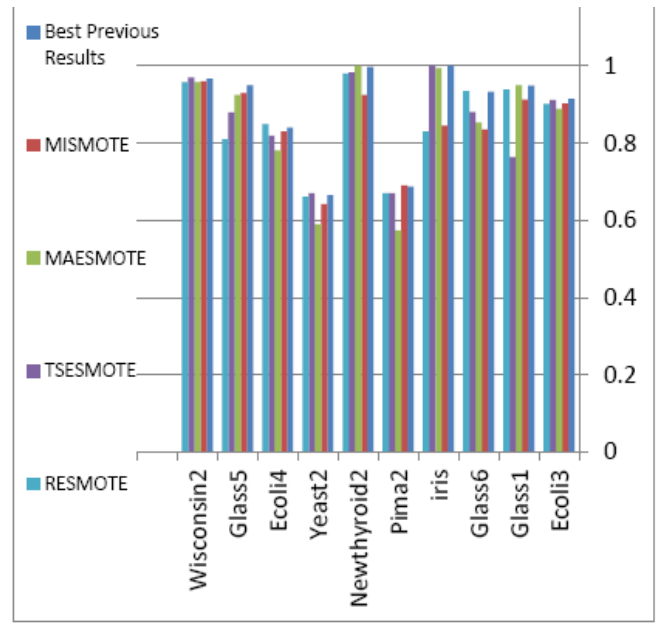

Fig.7 comparison of the best previous methods with our four proposed methods combined with 1NN classifier

In Fig.8, we briefly compare the best previous methods and our four proposed methods applying J48. As it is clear, with Ecoli2 and Pima2, MISMOTE has a bit better performance in comparison to the previous one. Unexpectedly, with Ecoli3 and Glass5, previous results indicate their best performance. Glass1 experiences its best performance by MAESMOTE. Meanwhile, TESMOTE has equal performance with Iris in comparison to previous methods. Moreover, New-Thyroid2 has its maximum value with MAESMOTE. Yeast2 with TSESMOTE has the best performance. Finally, Ecoli4 experiences its maximum score by RESMOTE.

One of the most important and comprehensible observation is that the results of the proposed methods with different datasets are among the best performing methods. Therefore, it appears that the use of these four methods as a sort of oversampling the minority class which leads to balance the datasets efficiently.

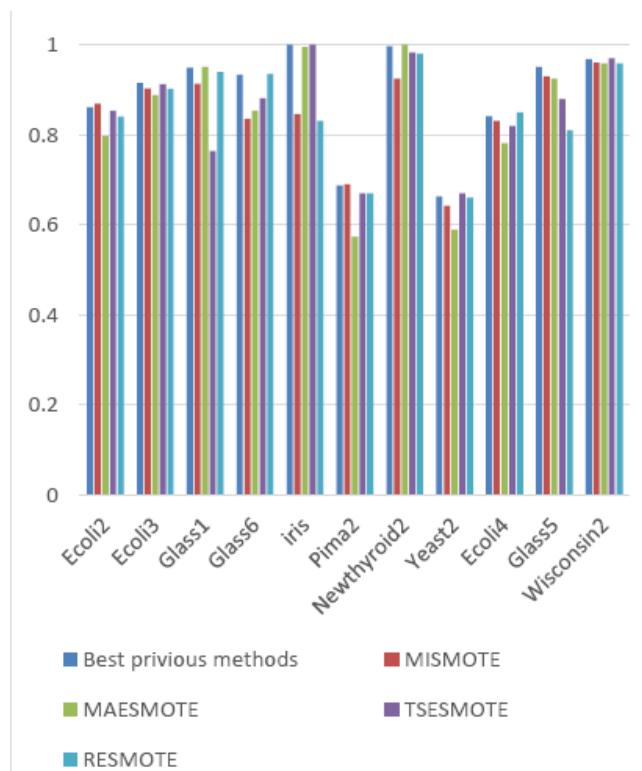

*Fig. 8. comparison of the best previous methods with our four proposed methods with J48 classifier*

## V. CONCLUSION

This paper has proposed four new SMOTE methods considered as pre-processing phase and their performance is evaluated by different classification algorithms. Firstly, our first proposed method which is enhanced SMOTE which mixture of clustering and classification techniques, dynamic selection, distance weighted and attribute weighted which is done by mutual information is tested on Tehran-Bazargan dataset with IR equal to 36. Then, a comparison between four proposed methods and WEKA's SMOTE is done on Tehran-Bazargan dataset. With different algorithms, we reach different improvements for instance, Decorate with MASMOTE, RESMOTE and TESMOTE has 0.02 percent improvement and reaches 1. Considering AdaboostM1 as the classification algorithm, TESMOTE, RESMOTE and MAESMOTE with 0.03 percent improvement reach 1. At the same time, Bagging with MAESMOTE, RESMOTE and TESMOTE has the same amount of improvement as MAESMOTE, RESMOTE and TESMOTE by J48 and all reach 1 with 0.02 percent improvement.

Then, different sorts of entropies are replaced by Mutual information in order to allocate weight to our features. Results were presented with different classification algorithms and almost in all of the datasets, an improvement is felt.

## Biographies

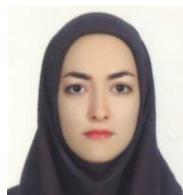

Sima Shaifirad received her MSc in Computer Science from Amirkabir University of Technology under supervision of Dr. Mehdi Ghatee in 2015. Her current research interests include data mining, imbalanced data and big data.

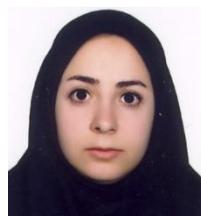

Azra Nazari received her MSc in Computer Science from Amirkabir University of Technology under supervision of Dr. Mehdi Ghatee in 2014. Her research interest covers big data, classification algorithms and imbalanced data.

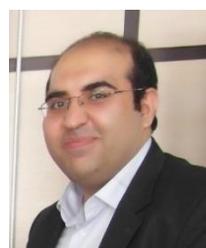

Mehdi Ghatee received his PhD in Computer Science from Amirkabir University of Technology under supervision of Prof. S. Mehdi Hashemi in 2009. He is currently the Chairman of the Department of Computer Science of Amirkabir University of Technology. His major is Soft Computing and Intelligent Transportation Systems. He has written more than 65 papers on national and international journals and conferences. He also has written a text book entitled as "Linear Optimization and Combinatorial Optimization". He is Project Manager of Nation Plan for Intelligent Transportation System Architecture in Iran.

Appendix:

Table 2. Details of distribution of the used datasets in experiments

| datasets | examples | IR | Positive class | Negative class |
|---|---|---|---|---|
| **Iran Accident** | 24872 | 36 | 676 | **24196** |
| Ecoli3 | 336 | 7 | 3 | **1,2,4,5,6,7,8** |
| Glass1 | 214 | 9 | 2 | **1,3,5,6,7** |
| Ecoli2 | 336 | 7 | 2 | **1,3,4,5,6,7,8** |
| Newthyroid2 | 215 | 5 | 2 | 1,3 |
| Glass6 | 184 | 9 | 6 | 1,2,7 |
| Pima2 | 768 | 8 | 2 | 1 |
| Iris | 150 | 4 | 1 | 2,3 |
| Yeast2 | 1484 | 2.46 | 2 | 1,3,4,5,6,7,8,9,10 |
| Ecoli4 | 336 | 8.60 | 4 | 1,2,3,5,6,7,8 |
| Glass5 | 214 | 15.46 | 5 | 1,2,3,6,7 |
| Wisconsin2 | **683** | **1.86** | **2** | 1 |

Table 3. Comparison of the previous methods and the proposed methods by 1NN

| | algorithms | datasets | | | | | | | | | | |
|---|---|---|---|---|---|---|---|---|---|---|---|---|
| | | Ecoli2 | Ecoli3 | Glass1 | Glass6 | iris | Pima2 | Newthyroid2 | Yeast2 | Ecoli4 | Glass5 | Wisconsin2 |
| **Previous methods** | imbalanced | 0.797 | 0.906 | 0.913 | 0.836 | **1.00** | 0.671 | 0.977 | 0.648 | 0.745 | 0.821 | 0.953 |
| | SMOTE | 0.836 | 0.907 | 0.923 | 0.883 | **1.00** | 0.673 | 0.992 | 0.652 | 0.821 | 0.866 | 0.961 |
| | B-SMOTE | 0.843 | 0.914 | 0.949 | 0.883 | **1.00** | 0.667 | 0.992 | 0.649 | 0.768 | 0.868 | 0.964 |
| | SL-SMOTE | 0.797 | 0.906 | 0.913 | 0.836 | **1.00** | 0.671 | 0.997 | 0.648 | 0.745 | 0.821 | 0.957 |
| | ROS | 0.797 | 0.906 | 0.913 | 0.836 | **1.00** | 0.671 | 0.997 | 0.648 | 0.745 | 0.821 | 0.953 |
| | AHC | 0.823 | 0.899 | 0.929 | 0.887 | **1.00** | 0.669 | 0.997 | 0.651 | 0.783 | 0.863 | 0.956 |
| | ADOMS | 0.843 | **0.915** | 0.942 | 0.836 | **1.00** | 0.674 | 0.989 | 0.665 | 0.841 | 0.809 | 0.959 |
| | ADASYN | 0.833 | 0.898 | 0.936 | 0.883 | **1.00** | 0.673 | 0.992 | 0.650 | 0.808 | 0.866 | 0.967 |
| | NCN-SMOTE | 0.845 | 0.893 | 0.929 | 0.933 | **1.00** | 0.687 | 0.986 | 0.657 | 0.821 | 0.923 | 0.963 |
| | GG-SMOTE | 0.861 | 0.896 | 0.942 | 0.883 | **1.00** | 0.688 | 0.989 | 0.663 | 0.834 | 0.95 | 0.966 |
| | RNG-SMOTE | 0.842 | 0.902 | 0.933 | 0.883 | **1.00** | 0.675 | 0.989 | 0.663 | 0.792 | 0.928 | 0.968 |
| **4 proposed pre-processing method** | MISMOTE | **0.869** | 0.903 | 0.913 | 0.835 | 0.845 | **0.690** | 0.925 | 0.642 | 0.830 | **0.930** | 0.960 |
| | MAESMOTE | 0.797 | 0.888 | **0.950** | 0.853 | 0.995 | 0.574 | **1.00** | 0.590 | 0.781 | 0.925 | 0.958 |
| | TSESMOTE | 0.853 | 0.912 | 0.764 | 0.881 | **1.00** | 0.670 | 0.983 | **0.670** | 0.820 | 0.880 | **0.970** |
| | RESMOTE | 0.840 | 0.902 | 0.939 | 0.935 | 0.830 | 0.670 | 0.980 | 0.661 | **0.850** | 0.810 | 0.958 |
| **MAX** | | 0.869 | 0.915 | 0.950 | 0.935 | 1.00 | 0.690 | 1.00 | 0.670 | 0.850 | 0.930 | 0.970 |
| **MIN** | | 0.797 | 0.888 | 0.764 | 0.835 | 0.830 | 0.574 | 0.980 | 0.648 | 0.745 | 0.821 | 0.953 |
| **AVE** | | 0.831 | 0.903 | 0.919 | 0.871 | 0.978 | 0.668 | 0.986 | 0.613 | 0.798 | 0.871 | 0.960 |

Table 4. Comparison of the previous methods and our four proposed methods by J48.

| | algorithms | datasets | | | | | | | | | | |
|---|---|---|---|---|---|---|---|---|---|---|---|---|
| | | Ecoli2 | Ecoli3 | Glass1 | Glass6 | iris | Pima2 | Newthyroid2 | Yeast2 | Ecoli4 | Glass5 | Wisconsin2 |
| **Previous methods** | imbalanced | 0.797 | 0.906 | 0.913 | 0.836 | **1.00** | 0.671 | 0.977 | 0.648 | 0.745 | 0.821 | 0.953 |
| | SMOTE | 0.836 | 0.907 | 0.923 | 0.883 | **1.00** | 0.673 | 0.992 | 0.652 | 0.821 | 0.866 | 0.961 |
| | B-SMOTE | 0.843 | 0.914 | 0.949 | 0.883 | **1.00** | 0.667 | 0.992 | 0.649 | 0.768 | 0.868 | 0.964 |
| | SL-SMOTE | 0.797 | 0.906 | 0.913 | 0.836 | **1.00** | 0.671 | 0.997 | 0.648 | 0.745 | 0.821 | 0.957 |
| | ROS | 0.797 | 0.906 | 0.913 | 0.836 | **1.00** | 0.671 | 0.997 | 0.648 | 0.745 | 0.821 | 0.953 |
| | AHC | 0.823 | 0.899 | 0.929 | 0.887 | **1.00** | 0.669 | 0.997 | 0.651 | 0.783 | 0.863 | 0.956 |
| | ADOMS | 0.843 | **0.915** | 0.942 | 0.836 | **1.00** | 0.674 | 0.989 | 0.665 | 0.841 | 0.809 | 0.959 |
| | ADASYN | 0.833 | 0.898 | 0.936 | 0.883 | **1.00** | 0.673 | 0.992 | 0.650 | 0.808 | 0.866 | 0.967 |
| | NCN-SMOTE | 0.845 | 0.893 | 0.929 | 0.933 | **1.00** | 0.687 | 0.986 | 0.657 | 0.821 | 0.923 | 0.963 |
| | GG-SMOTE | 0.861 | 0.896 | 0.942 | 0.883 | **1.00** | 0.688 | 0.989 | 0.663 | 0.834 | 0.95 | 0.966 |
| | RNG-SMOTE | 0.842 | 0.902 | 0.933 | 0.883 | **1.00** | 0.675 | 0.989 | 0.663 | 0.792 | 0.928 | 0.968 |
| **4 proposed pre-processing method** | MISMOTE | **0.869** | 0.903 | 0.913 | 0.835 | 0.845 | **0.690** | 0.925 | 0.642 | 0.830 | **0.930** | 0.960 |
| | MAESMOTE | 0.797 | 0.888 | **0.950** | 0.853 | 0.995 | 0.574 | **1.00** | 0.590 | 0.781 | 0.925 | 0.958 |
| | TSESMOTE | 0.853 | 0.912 | 0.764 | 0.881 | **1.00** | 0.670 | 0.983 | **0.670** | 0.820 | 0.880 | **0.970** |
| | RESMOTE | 0.840 | 0.902 | 0.939 | **0.935** | 0.830 | 0.670 | 0.980 | 0.661 | **0.850** | 0.810 | 0.958 |
| **MAX** | | 0.869 | 0.915 | 0.950 | 0.935 | 1.00 | 0.690 | 1.00 | 0.670 | 0.850 | 0.930 | 0.970 |
| **MIN** | | 0.797 | 0.888 | 0.764 | 0.835 | 0.830 | 0.574 | 0.980 | 0.648 | 0.745 | 0.821 | 0.953 |
| **AVE** | | 0.831 | 0.903 | 0.919 | 0.871 | 0.978 | 0.668 | 0.986 | 0.613 | 0.798 | 0.871 | 0.960 |